\documentclass[10pt,twocolumn,letterpaper]{article}

\usepackage{iccv}
\usepackage{times}
\usepackage{epsfig}
\usepackage{graphicx}
\usepackage{amsmath}
\usepackage{amssymb}

\usepackage{booktabs}
\usepackage{multirow}
\usepackage{comment}
\usepackage{mathtools}
\usepackage{cuted}
\usepackage[export]{adjustbox}

\usepackage[table]{xcolor}

\usepackage{etoolbox}
\newtoggle{comments}
\togglefalse{comments}
\iftoggle{comments}{%
    \newcommand{\deva}[1]{{\leavevmode\color{blue}[Deva: #1]}}
    \newcommand{\achal}[1]{{\leavevmode\color{orange}[Achal: #1]}}
    \newcommand{\pavel}[1]{{\leavevmode\color{red}[Pavel: #1]}}
}{%
    \newcommand{\deva}[1]{}
    \newcommand{\achal}[1]{}
    \newcommand{\pavel}[1]{}
}
\newcommand{\smallsec}[1]{\vspace{0.0em}\textbf{#1:}}
\newcommand{\davisnew}{DAVIS-Moving}
\newcommand{\ytvosmoving}{YTVOS-Moving}
\newcommand{\ytvosnew}{YTVOS-Moving}

\usepackage[font=small]{caption}
\DeclarePairedDelimiter\abs{\lvert}{\rvert}%

\usepackage[pagebackref=true,breaklinks=true,bookmarks=false]{hyperref}
\usepackage{cleveref}

\iccvfinalcopy %
\newtoggle{includeappendix}
\toggletrue{includeappendix}

\def\iccvPaperID{8}

\newcommand{\mytilde}{\raise.17ex\hbox{$\scriptstyle\mathtt{\sim}$}}

\ificcvfinal\fi
\begin{document}

\title{Towards Segmenting Anything That Moves}

\author{Achal Dave\\
Carnegie Mellon University\\
{\tt\small achald@cmu.edu}
\and
Pavel Tokmakov\\
Carnegie Mellon University\\
{\tt\small ptokmako@cmu.edu}
\and
Deva Ramanan\\
Carnegie Mellon University\\
{\tt\small deva@cmu.edu}}

\hbadness=99999

\maketitle

\begin{abstract}
    Detecting and segmenting individual objects, regardless of their category, is crucial for many applications such as action detection or robotic interaction. While this problem has been well-studied under the classic formulation of spatio-temporal grouping, state-of-the-art approaches do not make use of learning-based methods. To bridge this gap, we propose a simple learning-based approach for spatio-temporal grouping. Our approach leverages motion cues from optical flow as a bottom-up signal for separating objects from each other. Motion cues are then combined with appearance cues that provide a generic \textit{objectness} prior for capturing the full extent of objects. We show that our approach outperforms all prior work on the benchmark FBMS dataset. One potential worry with learning-based methods is that they might overfit to the particular type of objects that they have been trained on. To address this concern, we propose two new benchmarks for generic, moving object detection, and show that our model matches top-down methods on common categories, while out-performing both top-down and bottom-up methods on never-before-seen categories.
\end{abstract}

\section{Introduction}

People have the remarkable ability to thrive in a staggeringly diverse world, frequently encountering things they have never seen before. Our approaches for machine perception, meanwhile, often remain trapped in a {\em closed world}, as in the case of object recognition, where approaches are designed to recognize and name one of $N$ pre-defined classes. But practical robot autonomy requires robust perception in the open-world: even a self-driving car must be able to detect never-before-seen obstacles and debris, regardless of what particular semantic {\em name} it happens to associate with.

In the computer vision community, open-world recognition is typically addressed from a machine-learning perspective such as zero-shot learning~\cite{socher2013zero} or open-set classification~\cite{scheirer2014probability}. We advocate a different approach that has its roots in classic vision: perceptual grouping.
Specifically, we wish to segment out {\em all} moving object instances in a video stream, including never-before-seen object categories. 
Defining the notion of a generic, never-before-seen object is notoriously challenging~\cite{alexe2010object}. We intentionally focus on {\em moving} objects so as to take advantage of the ``common fate'' principle of grouping: pixels that move together should tend to be grouped together into objects~\cite{palmer1999vision}. 

\begin{figure}[t!]
\centering
\includegraphics[width=\linewidth,clip=true,trim=0 1mm 0 0]{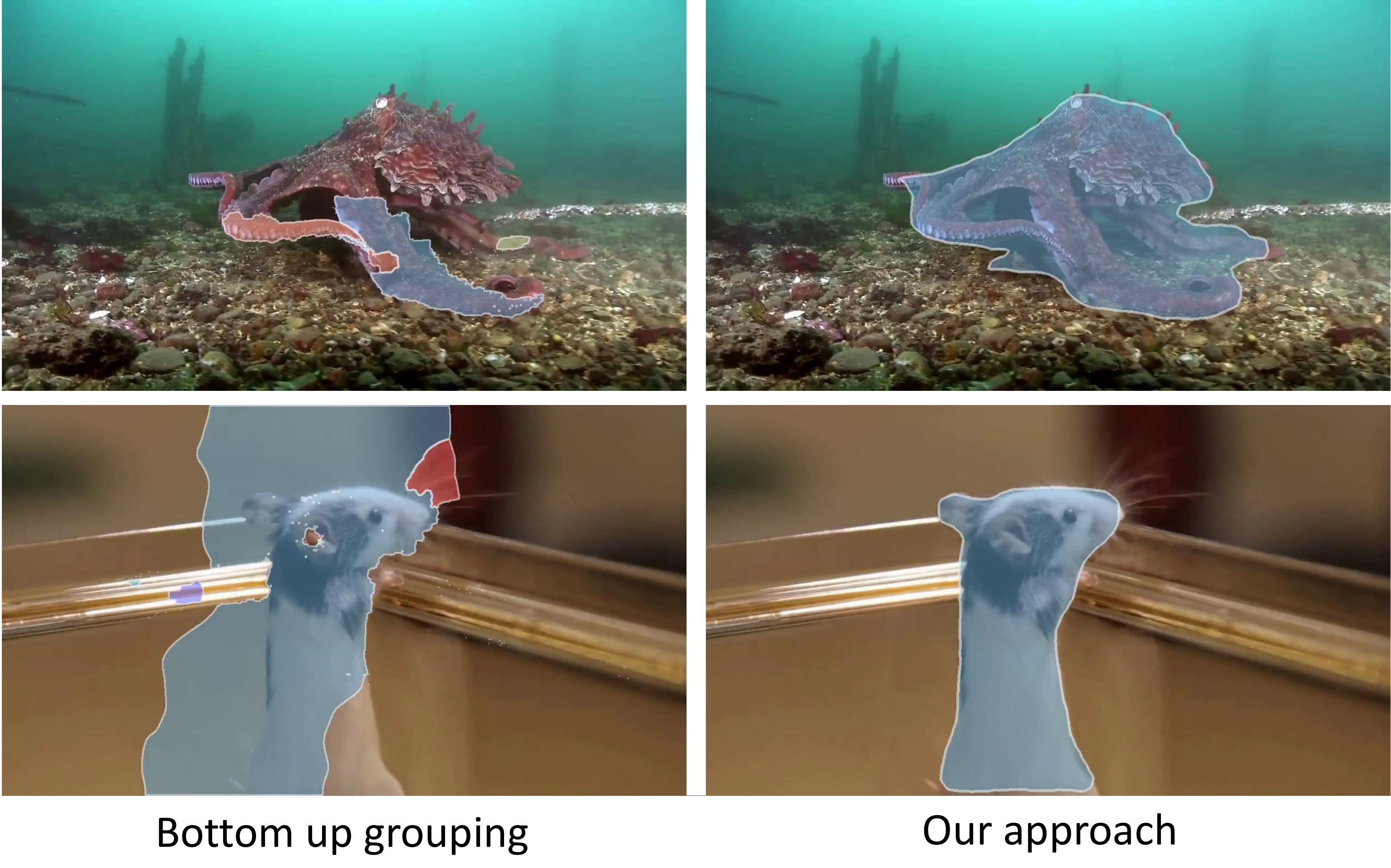}
   \caption{Detecting and segmenting all objects, regardless of category, is key for many perception and robotics tasks. Bottom-up grouping approaches, e.g.~\cite{keuper2015motion} (left), aim to tackle this task, but lag behind the quality of closed-world methods that detect a fixed set of $N$ categories. Our work (right) bridges this gap, accurately segmenting generic moving objects, even ones unseen in training.}
\label{fig:teaser}
\end{figure}

Indeed, the problem of spatio-temporal grouping is a classic ``mid-level''  visual understanding task, dating back to the iconic work of Marr~\cite{marr1982vision,wertheimer1923untersuchungen}. Pre-deep learning solutions tend to follow bottom-up computational strategies for self-organization and clustering, often of long-term pixel trajectories~\cite{ochs2014segmentation,keuper2015motion}. In the static image case, pixels can grouped by relying on Gestaltian notions of appearance similarity and curvilinear edge continuity~\cite{palmer1999vision}. One long-standing challenge in perceptual organization has been operationalizing these cues %
into an accurate algorithm for spatio-temporal grouping. {\em Our key observation is that many of the recent advances in closed-world instance segmentation can be repurposed for open-world spatio-temporal grouping}.

We first validate the performance of our proposed approach on the Freiburg Berkeley Motion Segmentation benchmark (FBMS). Because the standard measure used in FBMS does not penalize false positives, we find that trivial solutions can score well. We analyze the official metric in detail %
and propose a new, more informative evaluation. %
We achieve state-of-the-art results on both measures, and specifically outperform the next-best method of 
 Keuper et al.~\cite{keuper2015motion} by 11.4\% on our proposed measure.

To further study our method, we introduce the \davisnew{} and \ytvosnew{} benchmarks for motion-based grouping. We create these by selecting videos from the DAVIS 2017~\cite{Pont-Tuset_arXiv_2017} and YTVOS~\cite{xu2018youtube} datasets where {\em all} moving objects are labeled. On these new benchmarks, we strongly outperform top-down, closed world methods such as Mask R-CNN, as well as traditional bottom-up grouping methods. In particular, our approach is competitive with a top-down method for categories seen during training, but outperforms both top-down and bottom-up approaches for unseen categories.

To sum up, our contributions are three-fold: (1) we propose the first deep learning-based method for spatio-temporal grouping; (2) we propose a more informative metric and larger, more diverse benchmarks to enable further progress; (3) we report state-of-the-art results on the FBMS dataset and our larger, proposed benchmarks. The code and trained models will be made publicly available.

\section{Related Work}%
\label{sec:related_work}
\smallsec{Spatio-temporal grouping} Segmenting and tracking objects based on their motion has a rich history. An early work~\cite{shi1998motion} proposed treating this task as a spatio-temporal grouping problem, a philosophy espoused by a number of more recent approaches, including~\cite{grundmann2010efficient,brox2010object,keuper2015motion}, as well as~\cite{ochs2014segmentation}, which introduced FBMS. In particular, these methods track each pixel individually with optical flow, encode the motion information of a pixel in a compact descriptor and then obtain an instance segmentation by clustering the pixels based on motion similarity. Unlike these works, our approach is driven primarily by a top-down learning algorithm followed by a simple linking step to generate spatio-temporal segmentations. The most relevant approach in this respect is~\cite{fragkiadaki2015learning}, which trains a CNN to detect (but not segment) moving objects, and combines these detections with clustered pixel trajectories to derive segmentations.  By contrast, our approach directly outputs segmentations at each frame, which we link together with an efficient tracker. Very recently, Bideau \etal~\cite{bideau2018best} proposed to combine a heuristic-based motion segmentation method~\cite{narayana2013coherent,bideau2016s} with a CNN trained for semantic segmentation for the task of moving object segmentation. Their method, however, does not handle discontinuous motion. In addition, the fact that they rely strongly on heuristic motion estimates allows our learning-based approach to outperform their method on FBMS by a wide margin. In very recent work, Xie \etal \cite{xie2018object} introduced a deep learning approach for motion segmentation that segments and tracks moving objects using a recurrent neural network. By comparison, our method uses a simple, overlap-based tracker that performs competitively with the learned tracker from~\cite{xie2018object} while producing significantly fewer false positive segmentations (see Supplementary).

\smallsec{Foreground/Background Video Segmentation} Several works have focused on the binary version of the video segmentation task, separating all the moving objects from the background. Early approaches~\cite{faktor2014video,papazoglou2013fast,wang2015saliency,lee2011key} relied on heuristics in the optical flow field, such as closed motion boundaries in~\cite{papazoglou2013fast} to identified moving objects. These initial estimates were then refined with appearance, utilizing external cues, such as saliency maps~\cite{wang2015saliency}, or object shape estimates~\cite{lee2011key}. Another line of work focused on building probabilistic models of moving objects using optical flow orientations~\cite{narayana2013coherent,bideau2016s}. None of these methods are based on a robust learning framework and struggle to generalize well to unseen videos.  The recent introduction of a standard benchmark, DAVIS 2016~\cite{perazzi2016benchmark}, has led to a renewed interest. More recent approaches propose deep models for directly estimating motion masks, as in~\cite{jain2017fusionseg,tokmakov2017learning,tokmakov2018learning}. These approaches are similar to ours in that they also use a two-stream architecture to separately process motion and appearance, but they are unable to segment {\em individual} object instances, one of our primary goals. Our method separately segments and tracks each individual moving object in a video.

\smallsec{Object Detection} The task of segmenting object instances from still images has seen immense success in recent years, bolstered by large, standard datasets such as COCO~\cite{lin2014microsoft}. However, this standard task focuses on segmenting every instance of objects belonging to a fixed list of categories, leading to methods that are designed to be blind to objects that fall outside the categories in the training set.

Two recent works have focused on extending these models to detect generic objects.~\cite{hu2018learning} aims to generalize segmentation models to new categories, but requires bounding box annotations for each new category. More relevant to our approach,~\cite{jain2017pixel} aims to detect all ``object''-like regions in an image, outputting a binary objectness mask. While we share their goal of segmenting unseen objects, our approach additionally provides instance masks for each object.

\section{Approach}
\label{sec:approach}

We propose a two-stream spatio-temporal grouping method that uses appearance and motion cues to segment all moving objects in a video. Our approach, illustrated in \Cref{fig:two-stream}, takes a frame together with a corresponding optical flow as input, and passes them through an ``appearance stream'' (top) and a ``motion stream'' (bottom) respectively. The resulting features are combined and passed to the joint region proposal network (RPN), which learn to detect and segment moving objects irrespective of their category. 

Our approach shares inspiration with prior work that proposes two-stream approaches for object detection~\cite{gu2017ava,peng2016multi,gkioxari2015finding,fragkiadaki2015learning}, with two key differences. First, we design a novel region proposal module that learns to fuse both appearance and motion information to generate moving object detections. Second, to overcome the dearth of appropriate training data, we develop a stage-wise training strategy that allows us to leverage synthetic data to train our motion stream, image datasets to train our appearance stream, and a small amount of real video data to train the joint model. 

We first discuss the architecture and training strategy for the motion and appearance streams individually, and then detail how to combine these streams into one coherent architecture. Finally, we describe a simple tracker that we use for linking detections across time, allowing us to produce spatio-temporal groupings that span across many frames.

\subsection{Motion-based Segmentation}\label{sec:approach-motion}

\begin{figure}
    \centering
    \includegraphics[width=\linewidth]{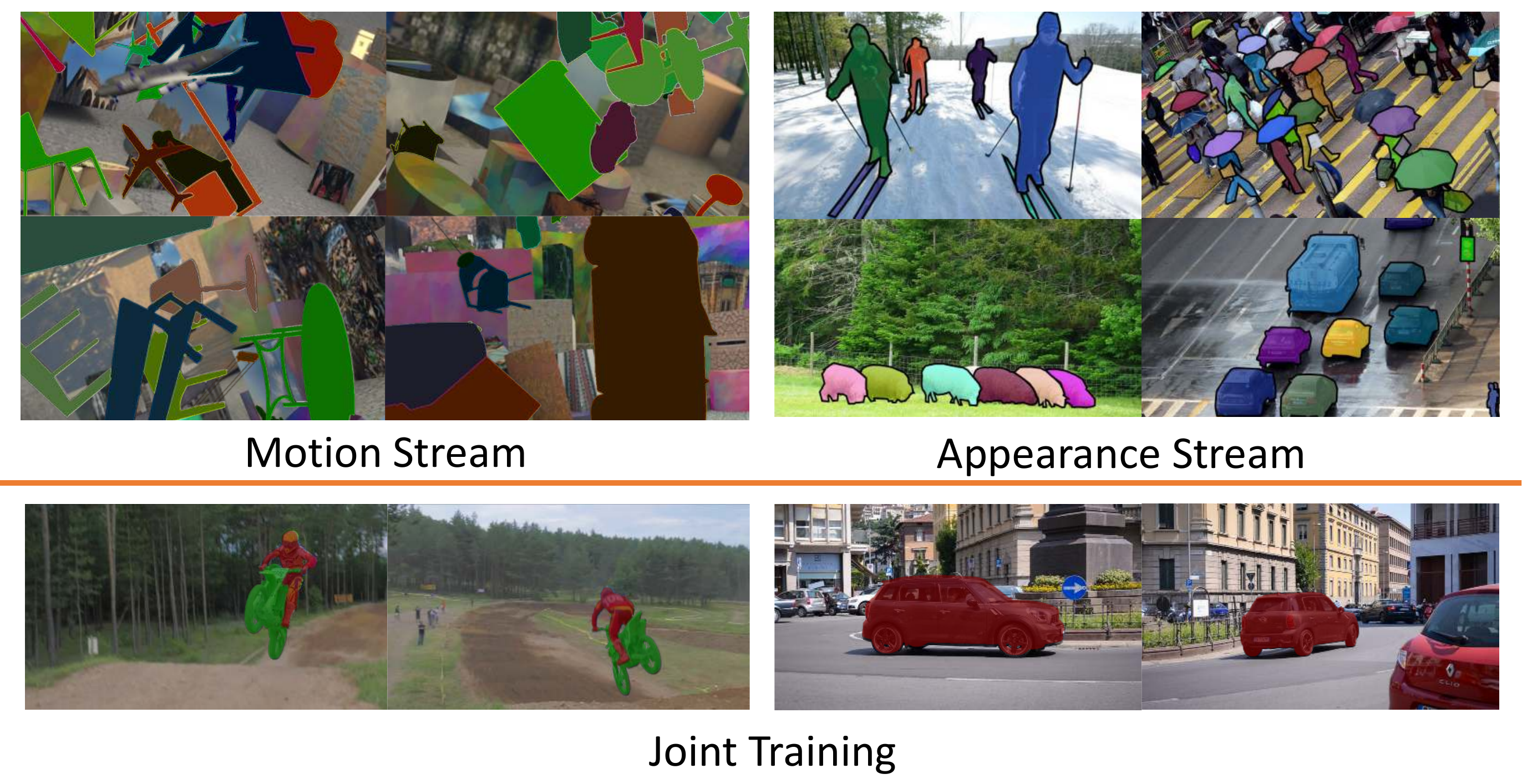}
    \caption{We train our motion stream on FlyingThings3D~\cite{MIFDB16} (top left), our appearance stream on COCO~\cite{lin2014microsoft} (top right), and our joint model on DAVIS'16~\cite{perazzi2016benchmark} and a YTVOS~\cite{xu2018youtube} subset (bottom).}
\label{fig:dataset-training}
\end{figure}

We start by training a motion-based instance segmentation model. As mentioned above, this requires videos with segmentation masks for all moving objects, which is difficult to obtain. Fortunately, prior work has shown that synthetic data can be used for some low-level tasks, such as flow estimation~\cite{dosovitskiy2015flownet} and binary motion segmentation~\cite{tokmakov2017learning}. Inspired by this, we train our motion stream on the FlyingThings3D dataset~\cite{MIFDB16}, which contains nearly 2,700 synthetically generated sequences of 3D objects traveling in randomized trajectories, captured with a camera also traveling along a random trajectory. The dataset provides groundtruth optical flow, as well as segmentations for both static and moving objects (See \Cref{fig:dataset-training}). We train our motion-stream using the moving instance labels from~\cite{tokmakov2017learning}, treating all moving objects as a single category, and all other pixels, including static objects, as background. The resulting model learns to segment moving objects irrespective of their category. In fact, this model is oblivious to the whole notion of an object and is capable of segmenting parts that exhibit independent motion (see Figure~\ref{fig:ablation-flow-qualitative}). 
We discuss more details and variants of this approach in \Cref{sec:ablation-motion}.

\subsection{Appearance-based Segmentation}\label{sec:approach-appearance}
\begin{figure*}[t]
\centering
\includegraphics[width=0.7\linewidth]{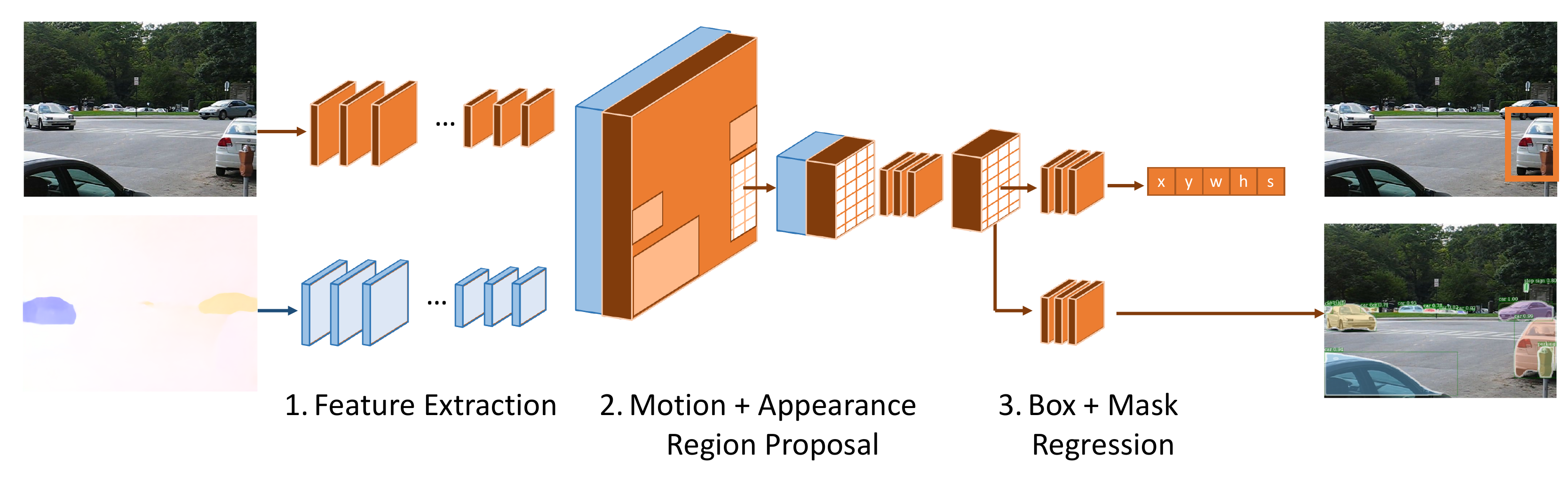}
   \caption{Our model uses an appearance stream (blue) and a motion stream (orange) to extract features from RGB and optical flow frames, respectively. Our region proposal network fuses features from both streams and passes them to the box and mask regression heads.}
\label{fig:two-stream}
\end{figure*}
In order to incorporate appearance information, we next train an image-based object segmentation model that aims to segment the full extent of generic objects. Fortunately, large datasets exist for training image-based instance segmentation models. Here, we train on the MS COCO dataset~\cite{lin2014microsoft}, which contains approximately 120,000 training images with instance segmentation masks for each object in 80 categories. We could train our appearance stream following the standard Mask R-CNN training procedure, which jointly localizes and classifies each object in an image belonging to the 80 categories. However, this results in a model that, while proficient at segmenting 80 categories, is blind to objects from any other, novel category. Instead, we train an ``objectness'' Mask R-CNN by combining each of the 80 categories into a single ``object'' category.  In \Cref{sec:ablation-appearance}, we will show that this ``objectness'' training (1) provides a significant improvement over standard training, and (2) leads to a model that generalizes surprisingly well to objects that are not labeled in MS COCO.

\subsection{Two-Stream Model}\label{sec:approach-two-stream}
Equipped with the individual appearance and motion streams, we now propose a two-stream architecture for fusing these information sources.
In order to clearly describe our two-stream model, we take a brief detour to describe the Mask R-CNN architecture. Mask R-CNN contains three stages: (1) \textbf{Feature extraction:} a ``backbone'' network, such as ResNet~\cite{he2016deep}, is used to extract features from an image. (2) \textbf{Region proposal:} A region proposal layer uses these features to selects regions likely to contain an object. Finally, (3) \textbf{Regression:} for each proposed region, the corresponding backbone features are pooled to a fixed size, and fed as input to bounding box and mask regression heads.

To build a two-stream instance segmentation model, we extract the backbone from our individual appearance-based and motion-based segmentation models. Next, as depicted in \Cref{fig:two-stream}, we propose a ``two-stream'' RPN that uses these two backbones, instead of a single backbone, to predict proposals from \textit{spatio-temporal} features, extracted from the optical flow (blue) and RGB (orange) backbones. These features are concatenated and fed to a short series of convolutional layers to reduce the dimensionality to match that of Mask R-CNN, allowing us to maintain the architecture of stages (2) and (3). Intuitively, we expect the appearance stream to behave as a generic object detector, and our motion stream to help detect novel objects that the appearance stream may miss and filter out static objects.

Although this may appear similar to prior approaches for building a two-stream detection model, it differs in a key detail: prior approaches obtain region proposals either only from appearance features~\cite{gu2017ava,gkioxari2015finding,fragkiadaki2015learning}, or from appearance and motion features individually~\cite{peng2016multi}. By contrast, we propose a novel proposal module that \textit{learns} to fuse motion and appearance features to find object-like regions.

We train our joint model on subsets of the DAVIS and YouTube Video Object Segmentation datasets (as detailed in \Cref{sec:experiments-dataset}). We experiment with various strategies for training this joint model in \Cref{sec:ablation-joint}.

\subsection{Tracking}\label{sec:approach-tracking}

So far, we have focused on segmenting moving objects in each frame of a video. To maintain object identities and to continue segmenting objects after they stop moving, we implement a simple, overlap-based tracker inspired by~\cite{Bewley2016_sort}. First, we remove all detections with score below $\alpha_{\text{low}}$. On the first frame, all high scoring detections (score $> \alpha_{\text{high}}$) are used to initialize a track, which we define simply as a sequence of linked detections. At each successive frame, we compute the mask intersection over union between the most recent segmentation for each active track and predicted objects at $t+1$, and use Hungarian Matching to assign predicted objects to tracks. Unmatched predictions are discarded if their score is $< \alpha_{\text{high}}$; else, they are used to initialize a new track. Tracks that have not been assigned a new object for up to $t_{\text{inactive}}$ frames are marked as inactive.

\smallsec{Tracking static objects} To continue tracking moving objects when they stop moving, we need to be able to detect static objects. A na\"{\i}ve way to do this is to run the objectness model trained in \Cref{sec:approach-appearance} in parallel with our two-stream model at every frame. However, this would be computationally expensive. Fortunately, our appearance stream shares the backbone of the objectness model. Thus, we only need to apply the (inexpensive) stages (2) and (3) of the objectness model on the appearance features extracted by our two-stream network. Using this, we can efficiently output a set of moving and static object predictions for each frame in a video. We merge the two outputs by removing any predicted static object that overlaps with a predicted moving object. We use the same tracker described above, using only moving objects to initialize tracks.

\section{Evaluation}\label{sec:evaluation}

To evaluate methods for spatio-temporal grouping, we desire a metric that rewards segmenting and tracking moving objects, but penalizes the detection of static objects or background. While there has been a rich line of prior work related to our goal, standard metrics surprisingly do not satisfy these criterion. We propose a novel metric that does. %

The default metric in FBMS~\cite{ochs2014segmentation} was designed for grouping-based approaches, but does not penalize false positive predictions. Recently, Bideau \etal~\cite{bideau2016detailed} tackled this issue by measuring the difference between the number of groundtruth moving objects and the number of predicted moving objects ($\Delta$ Obj). However, this complicates method comparisons by relying on two separate metrics; instead, we propose a single and intuitive F-measure that evaluates a method's ability to detect all and only moving objects.

\begin{figure}
    \centering
    \includegraphics[width=\linewidth]{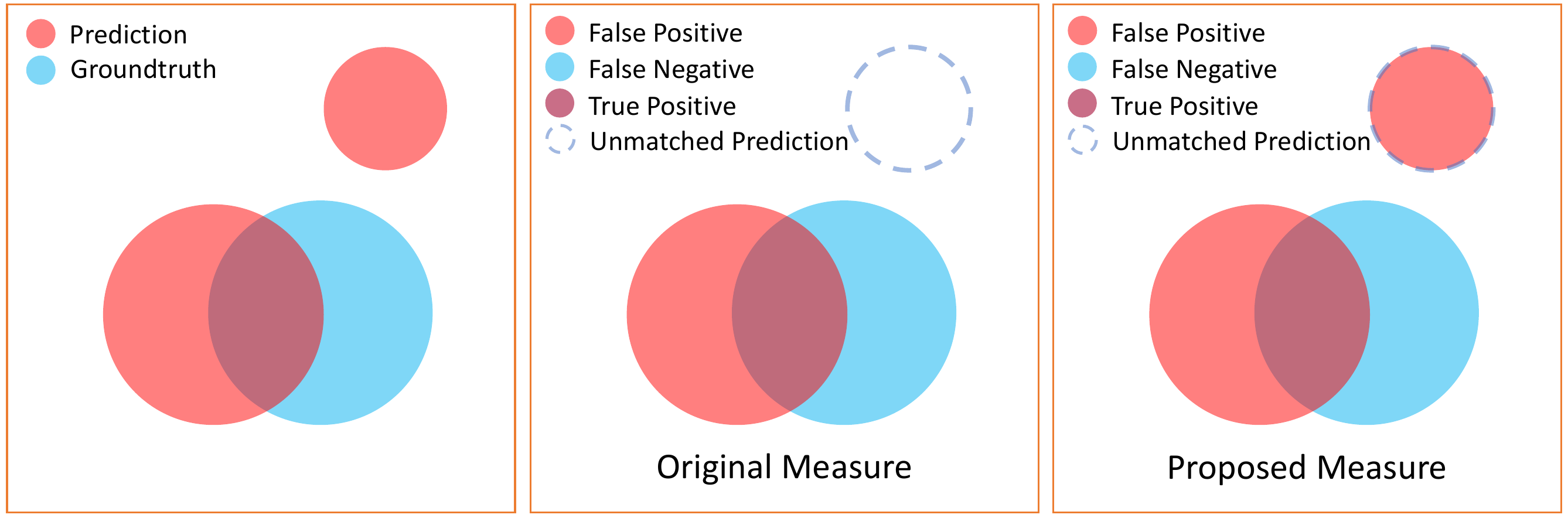}
    \caption{Left: we visualize a toy example with two predicted (red) segmentations and one groundtruth (blue) segmentation. While the original FBMS measure (middle) ignores predicted segments that do not match a groundtruth segment, such as the dashed circle, our proposed measure (right) penalizes all false-positives}
\label{fig:fbms-measure}
\end{figure}

\Cref{fig:fbms-measure} (middle) visualizes the default FBMS metric which matches each predicted segment with a groundtruth segment so as to maximize IoU overlap, {\em ignoring} any unmatched predictions. This means the default F-measure does not penalize false positive segments, unfairly favoring methods that generate a large number of predictions.
By contrast, our proposed F-measure, depicted in \Cref{fig:fbms-measure} (right), counts unmatched predictions as false positives.

More precisely, we describe our metric roughly following the notation in~\cite{ochs2014segmentation}. For each video, let $c_i$ be the pixels belonging to a predicted region $i$, and $g_j$ be all the pixels belonging to a groundtruth non-background region $j$. While~\cite{ochs2014segmentation} omits unlabeled pixels from evaluation, we include all pixels in the groundtruth.

Let $P_{ij}$ be the precision, $R_{ij}$ be the recall, and $F_{ij}$ be the F-measure corresponding to this pair of predicted and groundtruth regions, as follows:
\begin{align*}
P_{ij} = \frac{\abs{c_i \cap g_j}}{\abs{c_i}},
R_{ij} = \frac{\abs{c_i \cap g_j}}{\abs{g_j}},
F_{ij} = \frac{2 P_{ij}R_{ij}}{P_{ij} + R_{ij}}
\end{align*}

Following~\cite{ochs2014segmentation}, we use the Hungarian algorithm to find a matching between predictions and groundtruth that maximizes the sum of the F-measure over all assignments. Let $g(c_i)$ be the groundtruth matched to each predicted region; for any $c_i$ that is not matched to a groundtruth cluster, $g(c_i)$ is set to an empty region. We define our metric as follows:
\begin{align*}
P = \frac{\sum_i \abs{c_i \cap g(c_i)}}{\sum_j \abs{c_i}},
R = \frac{\sum_i \abs{c_i \cap g(c_i)}}{\sum_i \abs{g_j}},
F = \frac{2 PR}{P + R}
\end{align*}

Any unlabeled pixel in a predicted region $c_i$ will reduce precision and F-measure, penalizing the segmentation of static or unlabeled objects. In our experiments, we report results with both the official and our proposed measure.%

\section{Experiments}\label{sec:experiments}

We first analyze each component of our proposed model with experimental results. Next, we compare our approach to prior work in spatio-temporal grouping on three datasets.

\subsection{Datasets}\label{sec:experiments-dataset}
An ideal dataset for training our model would contain a large number of videos where every moving object has labeled instance masks, and static objects are not labeled. Three candidate datasets exist for this task: YouTube Video Object Segmentation (YTVOS)~\cite{xu2018youtube}, DAVIS 2016~\cite{perazzi2016benchmark}, and FBMS~\cite{ochs2014segmentation}. While YTVOS contains over 3,000 short videos with instance segmentation labels, not all objects in these videos are necessarily labeled, and both moving as well as static objects may be labeled. The DAVIS 2016 dataset contains instance segmentation masks (provided with DAVIS 2017) for only the moving objects, but only contains 30 training videos. Finally, although FBMS contains a total of 59 sequences with labeled instance segmentation masks for moving objects, prior work evaluates on the entire dataset, preventing us from training on any sequences in the dataset in order to provide a fair comparison.

To overcome this lack of data, we use heterogeneous data sources to train our model in a stagewise fashion. As described earlier, we train our appearance stream on COCO~\cite{lin2014microsoft}. We train our motion stream on FlyingThings3D~\cite{MIFDB16}, a synthetic dataset of 2,700 videos of randomly moving 3D objects. Finally, we fine-tune our joint model on DAVIS2016 and the training subset of \ytvosnew{}. We use a held-out set of 100 \ytvosnew{} sequences for evaluation.

\subsection{Implementation Details}\label{sec:experiments-implementation}
\smallsec{Network Architecture} Our two-stream model is built off Mask R-CNN~\cite{he2017mask} with a ResNet-50 backbone. We will publicly release the code and exact configuration for training, highlight some important details here, and note further details in supplementary. All our models are trained using the publicly available PyTorch implementation of Detectron~\cite{DetectronPytorch}. In general, we use the original hyper-parameters provided by the authors of Mask R-CNN. The backbone for every model is pre-trained on ImageNet~\cite{russakovsky2015imagenet}. When constructing our two-stream model, we initialize the bounding box and mask heads from the appearance-only model.

\smallsec{Tracking} We set the confidence threshold for initializing tracks, as described in \Cref{sec:approach-tracking}, to $\alpha_{\text{high}}=0.9$, and remove any detections with confidence lower than $\alpha_{\text{low}}=0.7$. We allow tracks to stay alive for up to $t_{\text{inactive}}=10$ frames (approximately $0.33s$ for most videos), although we found the final results are fairly insensitive to this parameter. To detect objects before they move, we first run our tracker forwards, and then backwards in time.

\subsection{Ablation analysis}\label{sec:ablation}
\smallsec{Evaluation} We analyze our model by benchmarking various configurations on the DAVIS 2016 dataset~\cite{perazzi2016benchmark}. For ablation, we found it helpful to use the standard detection mean average precision (mAP) metric~\cite{lin2014microsoft} in place of video object segmentation metrics, which require tracking and obfuscate analysis of our architecture choices. We report both detection and segmentation mAP at an IoU threshold of 0.5

\subsubsection{Motion stream}\label{sec:ablation-motion}

To begin, we explore training strategies for the motion stream of our model. We train our motion stream on the FlyingThings3D dataset, as described in \Cref{sec:approach-motion}. This dataset provides groundtruth flow, which we could use for training. However, at inference time, we only have access to noisy, estimated flow. In order to match flow in the real world, we estimate flow on FlyingThings3D using two optical flow estimation methods: FlowNet2 and LiteFlowNet. For both methods, we use the version of their model that is trained on synthetic data and fine-tuned on real data.

In~\Cref{tab:ablation-flow}, we compare three strategies for training on FlyingThings3D. We start by training using only FlowNet2 flow as input (``FlowNet2''). We hypothesize that training directly on noisy, estimated flow can lead to difficulties in early training. To overcome this, we train a variant starting with groundtruth flow, and fine-tune on FlowNet2 flow (``FlowNet2 $\leftarrow$ Groundtruth'' row). We find that this provides a significant improvement ($2.7\%$). We also considered using a more recent flow estimation method, LiteFlowNet~\cite{hui18liteflownet} (``LiteFlowNet $\leftarrow$ Groundtruth'' row). Surprisingly, we find that FlowNet2 provides significant improvements for detection, despite performing worse on standard flow estimation benchmarks. Qualitatively, we found that FlowNet2 provides sharper results along boundaries than LiteFlowNet, which may aid in localizing objects.

\begin{figure}
\centering
\includegraphics[width=0.49\linewidth]{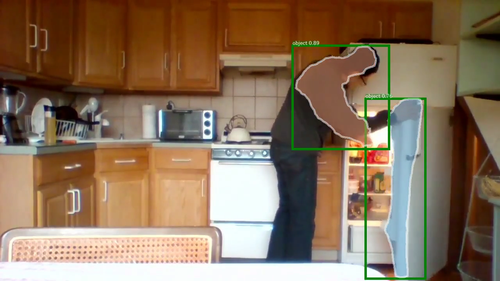}
\includegraphics[width=0.49\linewidth]{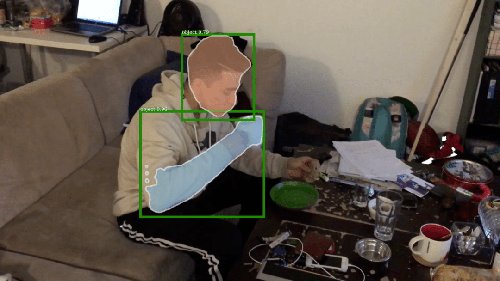}
   \caption[test]{Despite being trained for segmentation only on synthetic data, our motion stream (visualized) is able to separately segment object parts in real objects. See Sec.~\ref{sec:ablation-motion} for details.}
\label{fig:ablation-flow-qualitative}
\end{figure}

\Cref{fig:ablation-flow-qualitative} shows qualitative results of the motion stream. Despite never having seen real images with segmentation labels, this model is able to group together parts that move alike, while separating objects with disparate motion.

\begin{table}
\small
\centering
\begin{tabular}{lcc}
    \toprule
    Flow type & Det @ 0.5 & Seg @ 0.5 \\\midrule
    FlowNet2                             & 40.5  & 23.9  \\
    FlowNet2 $\leftarrow$ Groundtruth    & 43.2  & 24.1  \\
    LiteFlowNet $\leftarrow$ Groundtruth & 33.8  & 24.0  \\\bottomrule
\end{tabular}
\caption{Comparing training with different flow estimation methods on FlyingThings3D, reporting mAP on DAVIS '16 val. ``$\leftarrow$ Groundtruth'' means we first train with groundtruth (synthetic) flow. See \Cref{sec:ablation-motion} for details.}
\label{tab:ablation-flow}
\end{table}

\subsubsection{Appearance Stream}\label{sec:ablation-appearance}
While our motion stream is proficient at grouping similarly-moving pixels, it lacks any priors for real world objects and will not hesitate to oversegment common objects, such as the man in \Cref{fig:ablation-flow-qualitative}. To introduce these useful priors, we turn our attention to the appearance stream of our model.

As described in \Cref{sec:approach-appearance}, we train our appearance stream on the COCO dataset~\cite{lin2014microsoft}. We evaluate two variants of training. First, we train a standard, ``class-specific'' Mask R-CNN, that outputs a set of boxes and masks for each of the 80 categories in the COCO Dataset. At inference time, we combine the boxes and masks predicted for each category into a single ``object'' category. Second, we train an ``objectness'' Mask R-CNN, by collapsing all the categories in COCO to a single category \textit{before} training.

\begin{figure}
\centering
\includegraphics[width=\linewidth]{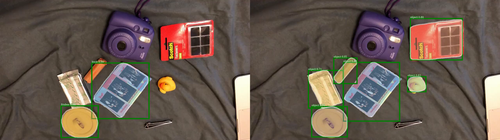}
   \caption{Unlike standard object detectors trained on COCO (left), our
   objectness model (right) detects objects from categories outside of COCO,
   such as the packet of film, a roll of quarters, a rubber duck, and a packet
   of fasteners. Both models visualized at confidence threshold of 0.7. See
   \Cref{sec:ablation-appearance} for details.}
\label{fig:ablation-appearance-objectness}
\end{figure}

We show results from these two variants in \Cref{tab:ablation-appearance}. Our ``objectness'' model significantly outperforms the standard ``class-specific'' model by nearly 8\%. We further compare the two models qualitatively in \Cref{fig:ablation-appearance-objectness}, noting that our objectness model better generalizes to non-COCO categories.

\begin{table}
\small
\centering
\begin{tabular}{lcc}\toprule
    COCO Training     & Det @ 0.5 & Seg @ 0.5 \\\midrule
    Class-specific & 42.0  & 40.2  \\
    Objectness     & 49.8  & 48.3  \\\bottomrule
\end{tabular}
\caption{Comparison of training our appearance stream with and without category labels on MS COCO (Class-specific and Objectness, respectively), reporting mAP on DAVIS '16 val. Training without category labels allows the model to generalize beyond the training categories. See also \Cref{fig:ablation-appearance-objectness} and \Cref{sec:ablation-appearance}}
\label{tab:ablation-appearance}
\end{table}

\subsubsection{Joint training}\label{sec:ablation-joint}
Finally, we combine our appearance and flow streams in a single two-stream model, depicted in~\Cref{fig:two-stream} and described in detail in~\Cref{sec:approach-two-stream}. We experiment with different strategies for training this joint model. Throughout these experiments, we initialize the flow stream with the ``FlowNet2 $\leftarrow$ Groundtruth'' model from \Cref{sec:ablation-motion}, and use the objectness model from \Cref{sec:ablation-appearance} to initialize the appearance stream, the box and mask prediction heads, and the RPN. We show the results in \Cref{tab:ablation-joint-freeze}.

We start by training this joint model directly on the DAVIS 2016 training set, which achieves 79.1\% mAP. We note that even with joint-training, using the objectness model for initialization provides a significant boost over using a category-specific detector (73.8\%). Next, to maintain the generalizability of the objectness model, we also train a variant where we freeze the weights of the appearance stream. This provides nearly a 3\% improvement in accuracy. Similarly, to maintain the generic ``grouping'' nature of the synthetically-trained flow stream, we freeze the flow stream, providing us with an additional 2\% improvement.

Finally, we hypothesize that while features from the flow stream are helpful for localizing generic moving objects, appearance information is sufficient for segmentation. We verify this hypothesis by training one last variant where the mask head uses only appearance stream features, and freeze its weights to those of the objectness model. Indeed, this provides a modest improvement of 1\% in segmentation AP.

\begin{table}
\small
\centering
\begin{tabular}{lcc}
    \toprule
    Variant & Det @ 0.5 & Seg @ 0.5 \\\midrule
    Joint Training, class-specific & 73.8  & 70.3  \\\midrule
    Joint Training, objectness     & 79.1  & 73.3  \\
    + Freeze appearance            & 81.9  & 76.7  \\
    + Freeze motion                & 83.7  & 76.4  \\
    + Freeze mask                  & 83.9  & 77.4  \\\bottomrule
\end{tabular}
\caption{Comparing two-stream training strategies, reporting mAP on DAVIS '16 val. Preserving knowledge from the individual streams is critical for good accuracy. See \Cref{sec:ablation-joint} for details.}
\label{tab:ablation-joint-freeze}
\end{table}

\smallsec{Training Data} Next, we train our joint model on YTVOS-Moving (\Cref{sec:experiments-dataset}) and show results in~\Cref{tab:ablation-joint-data}. Unfortunately, this dataset contains very few static objects, causing the model to detect both static and moving objects, leading to a significant (5\%) drop in performance. However, fine-tuning this model on the DAVIS 16 training set leads to our best model (DAVIS $\leftarrow$ YTVOS-moving).

\begin{table}
\small
\centering
\begin{tabular}{lcc}
\toprule
    Joint Training Data & Det @ 0.5 & Seg @ 0.5 \\\midrule
    DAVIS                           & 83.9  & 77.4  \\
    YTVOS-moving                    & 79.9  & 75.8  \\
    DAVIS $\leftarrow$ YTVOS-moving & 85.1  & 77.9  \\\bottomrule
\end{tabular}
\caption{Comparing training sources, reporting mAP on DAVIS '16 val. The lack of static objects in `YTVOS-moving' leads to worse performance, but fine-tuning on DAVIS provides the best model. See \Cref{sec:ablation-joint} for details.}
\label{tab:ablation-joint-data}
\end{table}

\subsection{Comparison to prior work}

\begin{figure}
    \centering
    \includegraphics[width=\linewidth]{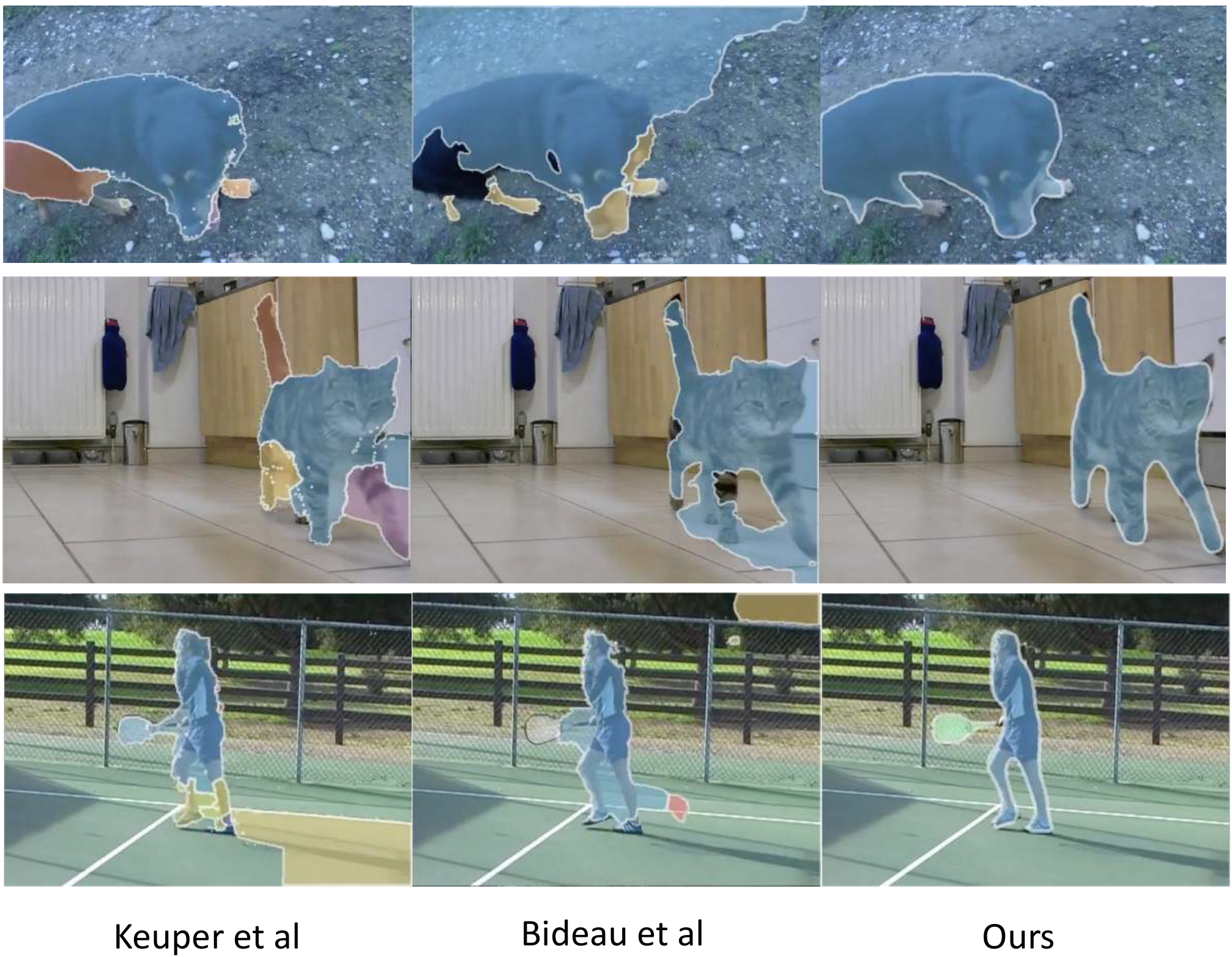}
    \caption{Qualitative results comparing our approach to two state-of-the-art methods. Prior work frequently exhibits over- or under-segmentation, such as the cat (middle row, ~\cite{keuper2015motion}) and the dog (top row, ~\cite{bideau2018best}), respectively. Our method fuses motion and appearance information to segment the full extent of moving objects.}
\label{fig:qualitative-final-compare}
\end{figure}

\smallsec{Official FBMS} We first evaluate our method against prior work on the standard FBMS benchmark in~\Cref{tab:fbms59_official}. As discussed in~\Cref{sec:experiments-dataset}, this metric does not penalize false positive detections. As expected, our appearance stream alone, despite segmenting both static and moving objects, performs best on this metric (`Ours-A'), outperforming all prior work by 6.4\% in F-measure on the TestSet, and 2.2\% on the TrainingSet~\footnote{Note that despite the name, we do not use either set for training.}. For completion, we also report the performance of our joint model (`Ours-J'), which compares favorably to state-of-the-art despite the flawed metric. Our improvements on this metric are likely driven by improvements in segmentation boundaries (see \Cref{fig:qualitative-final-compare}).

\begin{table}[bt]
\centering
\resizebox{\linewidth}{!}{
    \begin{tabular}{lc@{\hspace{1em}}c@{\hspace{1em}}c@{\hspace{1em}}c@{\hspace{1em}}c@{\hspace{1em}}c@{\hspace{1em}}c@{\hspace{1em}}c}
    \toprule
      & \multicolumn{4}{c}{Training set} &
          \multicolumn{4}{c}{Test set}\\
        & P           & R           & F           & $N/65$
        & P           & R           & F           & $N/69$ \\\midrule
    \cite{taylor2015causal}
        & 83.0        & 70.1        & 76.0        & 23
        & 77.9        & 59.1        & 67.3        & 15 \\
    \cite{keuper2015motion}
        & 86.9        & 71.3        & 78.4        & 25
        & 87.6        & 70.2        & 77.9        & 25 \\
    \cite{yang2015self}
        & 89.5        & 70.7        & 79.0        & 26
        & 91.5        & 64.8        & 75.8        & 27 \\
    \cite{khan2017coarse}
        & {\bf 93.0}  & 72.7        & 81.6        & 29
        & {\bf 95.9}  & 65.5        & 77.9        & 28 \\ \midrule
    Ours-A
        & 89.2        & {\bf 79.0}  & {\bf 83.8}  & {\bf 43}
        & 88.6        & {\bf 80.4}  & {\bf 84.3}  & {\bf 40}  \\
    Ours-J
        & 85.1        & 78.5        & 81.7        & 39
        & 80.8        & 75.8        & 78.2        & 39 \\\bottomrule
\end{tabular}
}
\caption{FBMS 59 results using the official metric~
\cite{ochs2014segmentation}, which does not penalize detecting unlabeled
objects. We report precision (P), recall (R), F-measure (F), and the number of
objects for which the F-measure $>0.75$ (N).  Ours-A is our model's appearance
stream only, and Ours-J is our joint model. Both Ours-A and Ours-J out-perform
all prior work. As expected, since this metric does not penalize false
positives, Ours-A outperforms Ours-J.}%
\label{tab:fbms59_official}
\end{table}

\smallsec{Proposed FBMS} Finally, we report results on our proposed metric in \Cref{tab:fbms59_custom}. Recall that our proposed metric generally follows the official metric, but additionally penalizes detection of static objects. We compare to all methods from \Cref{tab:fbms59_official} whose final results on FBMS were accessible or provided by the authors through personal communication. On this proposed metric, we first note that, as expected, the performance of our appearance model baseline is significantly worse than our final, joint model, by 9.2\% on TestSet and 6\% on TrainingSet in F-measure. More importantly, our final model strongly out performs prior work in F-measure by 11.3\% on the TestSet, and 6.1\% on the TrainingSet. In addition to improving segmentation boundaries, our approach effectively removes spurious segmentations of background regions and object parts (\Cref{fig:qualitative-final-compare}).

\begin{table}[bt]
\small
 \centering
    \begin{tabular}{lc@{\hspace{1em}}c@{\hspace{1em}}c@{\hspace{1.5em}}c@{\hspace{1em}}c@{\hspace{1em}}c@{\hspace{1em}}c}
      \toprule
      & \multicolumn{3}{c}{Training set} &
          \multicolumn{3}{c}{Test set}\\
        & P           & R           & F
        & P           & R           & F           \\\midrule
    \cite{taylor2015causal}
        & 74.8        & 61.7        & 65.5
        & 66.8        & 49.2        & 53.6       \\
    \cite{keuper2015motion}
        & 68.1        & 68.5        & 67.1
        & 70.0        & 64.6        & 65.0        \\
    Ours-A
        & 61.6        & {\bf 80.4}    & 64.0
        & 66.8        & {\bf 84.7}    & 70.3        \\
    Ours-J
        & {\bf 75.0}  & 77.8        & {\bf 73.2}
        & {\bf 77.0}  & 83.0        & {\bf 76.3}  \\\bottomrule
\end{tabular}
\caption{FBMS 59 results on our proposed metric. Ours-A is our appearance stream, Ours-J is our joint model.
We compare to prior methods for which we were able to obtain code or
results.}
\label{tab:fbms59_custom}
\end{table}

\smallsec{Qualitative results} We qualitatively compare our
approach with Keuper \etal~\cite{keuper2015motion} and Bideau
\etal~\cite{bideau2018best} in
\Cref{fig:qualitative-final-compare}\footnote{~\cite{bideau2018best}
only segments objects while they move. We provide an evaluation using an alternative FBMS
labeling they propose in our supplementary.} In the top
row of \Cref{fig:qualitative-final-compare},~\cite{keuper2015motion}
oversegments the dog into multiple parts, and~\cite{bideau2018best} merges the
dog with the background, whereas our approach fully segments
the dog.  Similarly, the cat in the middle row is over-segmented by
~\cite{keuper2015motion} and under-segmented by~\cite{bideau2018best}, but
well-segmented by our approach. In the final row, both~\cite{keuper2015motion}
and~\cite{bideau2018best} exhibit segmentation and tracking errors; the region
corresponding to the man's foot (colored yellow for Keuper \etal and red for
Bideau \etal) are mistakenly tracked into a background region thus segmenting
part of the background as a moving object. Meanwhile, our object-based tracker
fully segments the person and the tennis racket with high precision. We show
further qualitative results in supplementary material.

\smallsec{\davisnew{}} We further evaluate our method on a subset of the DAVIS 17 dataset. Unlike DAVIS 2016, the 2017 version provides instance-level masks for objects, but contains sequences with labeled static or unlabeled moving objects. For evaluation, we manually select 22 of 30 validation videos without these issues, and refer to this subset as \davisnew{}. We compare to~\cite{keuper2015motion}, the best FBMS method we can obtain code for, with our proposed metric in \Cref{tab:davis_proposed}. Surprisingly, we find a much larger gap in performance on this dataset; while~\cite{keuper2015motion} achieves 42.3\% on F-measure with our proposed metric, our approach improves significantly to 77.9\%. We believe this gap may be due to faster, more articulated motion and higher resolution videos in DAVIS 17, which severely affect~\cite{keuper2015motion} but not our method.

\begin{table}[bt]
\small
 \centering
    \begin{tabular}{lc@{\hspace{1em}}c@{\hspace{1em}}c@{\hspace{1.5em}}}
      \toprule
        & P           & R           & F           \\\midrule
    \cite{keuper2015motion}
        & 39.4        & 53.8        & 42.3        \\
    Mask R-CNN
        & 70.8        & 75.6         & 71.6 \\
    Ours-A & 67.7 & 77.1 & 70.9 \\
    Ours
        & {\bf 78.3}  & {\bf 78.8}   & {\bf 78.1}  \\\bottomrule
\end{tabular}
\caption{\davisnew{} results on our proposed metric. We compare to the best FBMS method for which we could obtain code.}
\label{tab:davis_proposed}
\end{table}

\begin{table}[bt]
 \small
 \centering
    \begin{tabular}{lc@{\hspace{1em}}c@{\hspace{1em}}}
      \toprule
        & \multicolumn{2}{c}{F} \\
        & Overall & Novel objects \\\midrule
    \cite{keuper2015motion}\footnotemark & 26.6 & 31.9      \\
    Mask R-CNN & 53.6  & 40.6      \\
    Ours-A w/o YTVOS & \textbf{69.0} & 65.8 \\
    Ours-J w/o YTVOS & 68.2 & \textbf{67.7} \\\bottomrule
\end{tabular}
\caption{\ytvosnew{} results on our proposed metric. For fairness, we evaluate our method \textit{without} YTVOS training. We compare to the best FBMS method for which we could obtain code.}
\label{tab:ytvos_proposed}
\end{table}
\footnotetext{\cite{keuper2015motion} errored on some sequences, so we report numbers on a subset. By comparison, Ours w/o YTVOS achieves 71.9\% F-measure on this subset.}

\smallsec{\ytvosnew}
Finally, we evaluate on sequences from \ytvosmoving{} (selected from YTVOS, as
described in \Cref{sec:experiments-dataset}). Unlike FBMS and DAVIS, YTVOS
contains diverse objects, such as octopuses and snakes. For fairness, we evaluate
a version of our final model that was never trained on YTVOS, and show
results in \Cref{tab:ytvos_proposed}.
In the first column, we report the F-measure overall on the \ytvosmoving{} dataset.
In the second column, we report F-measure on a subset of videos containing objects not from COCO categories, which our model has not seen during training.
Overall, we find that Mask R-CNN struggles to detect the diverse objects in this data.
We first show that our appearance stream alone (Ours-A) significantly improves over this baseline, from 53.6\% to 69.0\%.
Our final approach (Ours-J) performs comparably overall to the appearance stream alone, suggesting that it is a key driver of our improvements on this data.
On novel objects, however, our final approach provides modest improvements by incorporating the motion stream.
We show further qualitative results in the appendix.

\section{Conclusion}
We proposed a simple learning-based approach for spatio-temporal grouping. Our method provides two key insights. First, learning based approaches are able to generalize to never-before-seen objects (\Cref{sec:ablation-appearance}). Second, synthetic data can be used to train a truly generic grouping method with little priors on real world objects. As a result, our approach achieves state-of-the-art results on the FBMS benchmark dataset. Finally, to enable further research in this direction, we introduced a new metric as well as two new benchmarks (\davisnew{}, \ytvosnew{}).

\ificcvfinal{
\vspace{0.3em}
\noindent\smallsec{Acknowledgements}
We thank Pia Bideau for providing evaluation code, Nadine Chang, Kenneth Marino
and Senthil Purushwalkam for reviewing drafts and
discussions.  Supported by the Intelligence Advanced Research Projects Activity
(IARPA) via Department of Interior/Interior Business Center (DOI/IBC) contract
number D17PC00345. The U.S. Government is authorized to reproduce and distribute
reprints for Governmental purposes not withstanding any copyright annotation
theron. Disclaimer: The views and conclusions contained herein are those of the
authors and should not be interpreted as necessarily representing the official
policies or endorsements, either expressed or implied of IARPA, DOI/IBC or the
U.S. Government.}\fi

\clearpage
{\small
\bibliographystyle{ieee_fullname}
\bibliography{main}

\begin{thebibliography}{10}\itemsep=-1pt

\bibitem{alexe2010object}
Bogdan Alexe, Thomas Deselaers, and Vittorio Ferrari.
\newblock What is an object?
\newblock In {\em CVPR}, 2010.

\bibitem{Bewley2016_sort}
Alex Bewley, Zongyuan Ge, Lionel Ott, Fabio Ramos, and Ben Upcroft.
\newblock Simple online and realtime tracking.
\newblock In {\em ICIP}, 2016.

\bibitem{bideau2016detailed}
Pia Bideau and Erik Learned-Miller.
\newblock A detailed rubric for motion segmentation.
\newblock {\em arXiv preprint arXiv:1610.10033}, 2016.

\bibitem{bideau2016s}
Pia Bideau and Erik Learned-Miller.
\newblock It's moving! a probabilistic model for causal motion segmentation in
  moving camera videos.
\newblock In {\em ECCV}, 2016.

\bibitem{bideau2018best}
Pia Bideau, Aruni RoyChowdhury, Rakesh~R Menon, and Erik Learned-Miller.
\newblock The best of both worlds: Combining {CNN}s and geometric constraints
  for hierarchical motion segmentation.
\newblock In {\em CVPR}, 2018.

\bibitem{brox2010object}
Thomas Brox and Jitendra Malik.
\newblock Object segmentation by long term analysis of point trajectories.
\newblock In {\em ECCV}, 2010.

\bibitem{dosovitskiy2015flownet}
Alexey Dosovitskiy, Philipp Fischer, Eddy Ilg, Philip Hausser, Caner Hazirbas,
  Vladimir Golkov, Patrick Van Der~Smagt, Daniel Cremers, and Thomas Brox.
\newblock Flownet: Learning optical flow with convolutional networks.
\newblock In {\em ICCV}, 2015.

\bibitem{faktor2014video}
Alon Faktor and Michal Irani.
\newblock Video segmentation by non-local consensus voting.
\newblock In {\em BMVC}, 2014.

\bibitem{fragkiadaki2015learning}
Katerina Fragkiadaki, Pablo Arbelaez, Panna Felsen, and Jitendra Malik.
\newblock Learning to segment moving objects in videos.
\newblock In {\em CVPR}, 2015.

\bibitem{gkioxari2015finding}
Georgia Gkioxari and Jitendra Malik.
\newblock Finding action tubes.
\newblock In {\em CVPR}, 2015.

\bibitem{grundmann2010efficient}
Matthias Grundmann, Vivek Kwatra, Mei Han, and Irfan Essa.
\newblock Efficient hierarchical graph-based video segmentation.
\newblock In {\em CVPR}, 2010.

\bibitem{gu2017ava}
Chunhui Gu, Chen Sun, Sudheendra Vijayanarasimhan, Caroline Pantofaru, David~A
  Ross, George Toderici, Yeqing Li, Susanna Ricco, Rahul Sukthankar, Cordelia
  Schmid, et~al.
\newblock {AVA}: A video dataset of spatio-temporally localized atomic visual
  actions.
\newblock In {\em CVPR}, 2017.

\bibitem{he2017mask}
Kaiming He, Georgia Gkioxari, Piotr Doll{\'a}r, and Ross Girshick.
\newblock Mask {R-CNN}.
\newblock In {\em ICCV}, 2017.

\bibitem{he2016deep}
Kaiming He, Xiangyu Zhang, Shaoqing Ren, and Jian Sun.
\newblock Deep residual learning for image recognition.
\newblock In {\em CVPR}, 2016.

\bibitem{hu2018learning}
Ronghang Hu, Piotr Dollár, Kaiming He, Trevor Darrell, and Ross Girshick.
\newblock Learning to segment every thing.
\newblock In {\em CVPR}, 2018.

\bibitem{hui18liteflownet}
Tak-Wai Hui, Xiaoou Tang, and Chen~Change Loy.
\newblock {LiteFlowNet}: A lightweight convolutional neural network for optical
  flow estimation.
\newblock In {\em CVPR}, 2018.

\bibitem{ilg2017flownet}
Eddy Ilg, Nikolaus Mayer, Tonmoy Saikia, Margret Keuper, Alexey Dosovitskiy,
  and Thomas Brox.
\newblock Flownet 2.0: Evolution of optical flow estimation with deep networks.
\newblock In {\em CVPR}, 2017.

\bibitem{jain2017fusionseg}
Suyog~Dutt Jain, Bo Xiong, and Kristen Grauman.
\newblock Fusionseg: Learning to combine motion and appearance for fully
  automatic segmention of generic objects in videos.
\newblock In {\em CVPR}, 2017.

\bibitem{jain2017pixel}
Suyog~Dutt Jain, Bo Xiong, and Kristen Grauman.
\newblock Pixel objectness.
\newblock {\em arXiv preprint arXiv:1701.05349}, 2017.

\bibitem{keuper2015motion}
Margret Keuper, Bjoern Andres, and Thomas Brox.
\newblock Motion trajectory segmentation via minimum cost multicuts.
\newblock In {\em ICCV}, 2015.

\bibitem{khan2017coarse}
Naeemullah Khan, Byung-Woo Hong, Anthony Yezzi, and Ganesh Sundaramoorthi.
\newblock Coarse-to-fine segmentation with shape-tailored continuum scale
  spaces.
\newblock In {\em CVPR}, 2017.

\bibitem{lee2011key}
Yong~Jae Lee, Jaechul Kim, and Kristen Grauman.
\newblock Key-segments for video object segmentation.
\newblock In {\em ICCV}, 2011.

\bibitem{lin2014microsoft}
Tsung-Yi Lin, Michael Maire, Serge Belongie, James Hays, Pietro Perona, Deva
  Ramanan, Piotr Doll{\'a}r, and C~Lawrence Zitnick.
\newblock Microsoft coco: Common objects in context.
\newblock In {\em ECCV}. Springer, 2014.

\bibitem{marr1982vision}
David Marr.
\newblock Vision: A computational investigation into the human representation
  and processing of visual information. mit press.
\newblock {\em Cambridge, Massachusetts}, 1982.

\bibitem{MIFDB16}
N. Mayer, E. Ilg, P. H{\"a}usser, P. Fischer, D. Cremers, A. Dosovitskiy, and
  T. Brox.
\newblock A large dataset to train convolutional networks for disparity,
  optical flow, and scene flow estimation.
\newblock In {\em CVPR}, 2016.

\bibitem{narayana2013coherent}
Manjunath Narayana, Allen Hanson, and Erik Learned-Miller.
\newblock Coherent motion segmentation in moving camera videos using optical
  flow orientations.
\newblock In {\em ICCV}, 2013.

\bibitem{ochs2014segmentation}
Peter Ochs, Jitendra Malik, and Thomas Brox.
\newblock Segmentation of moving objects by long term video analysis.
\newblock {\em TPAMI}, 36(6):1187--1200, 2014.

\bibitem{palmer1999vision}
Stephen~E Palmer.
\newblock {\em Vision science: Photons to phenomenology}.
\newblock MIT press, 1999.

\bibitem{papazoglou2013fast}
Anestis Papazoglou and Vittorio Ferrari.
\newblock Fast object segmentation in unconstrained video.
\newblock In {\em ICCV}, 2013.

\bibitem{peng2016multi}
Xiaojiang Peng and Cordelia Schmid.
\newblock Multi-region two-stream {R-CNN} for action detection.
\newblock In {\em ECCV}. Springer, 2016.

\bibitem{perazzi2016benchmark}
Federico Perazzi, Jordi Pont-Tuset, Brian McWilliams, Luc Van~Gool, Markus
  Gross, and Alexander Sorkine-Hornung.
\newblock A benchmark dataset and evaluation methodology for video object
  segmentation.
\newblock In {\em CVPR}, 2016.

\bibitem{Pont-Tuset_arXiv_2017}
Jordi Pont-Tuset, Federico Perazzi, Sergi Caelles, Pablo Arbel\'aez, Alexander
  Sorkine-Hornung, and Luc {Van Gool}.
\newblock The 2017 {DAVIS} challenge on video object segmentation.
\newblock {\em arXiv:1704.00675}, 2017.

\bibitem{russakovsky2015imagenet}
Olga Russakovsky, Jia Deng, Hao Su, Jonathan Krause, Sanjeev Satheesh, Sean Ma,
  Zhiheng Huang, Andrej Karpathy, Aditya Khosla, Michael Bernstein, et~al.
\newblock {ImageNet} large scale visual recognition challenge.
\newblock {\em IJCV}, 115(3):211--252, 2015.

\bibitem{scheirer2014probability}
Walter~J Scheirer, Lalit~P Jain, and Terrance~E Boult.
\newblock Probability models for open set recognition.
\newblock {\em IEEE TPAMI}, 36(11):2317--2324, 2014.

\bibitem{shi1998motion}
Jianbo Shi and Jitendra Malik.
\newblock Motion segmentation and tracking using normalized cuts.
\newblock In {\em ICCV}, 1998.

\bibitem{siam2018video}
Mennatullah Siam, Chen Jiang, Steven Lu, Laura Petrich, Mahmoud Gamal, Mohamed
  Elhoseiny, and Martin Jagersand.
\newblock Video segmentation using teacher-student adaptation in a human robot
  interaction (hri) setting.
\newblock {\em ICRA}, 2019.

\bibitem{socher2013zero}
Richard Socher, Milind Ganjoo, Christopher~D Manning, and Andrew Ng.
\newblock Zero-shot learning through cross-modal transfer.
\newblock In {\em NIPS}, 2013.

\bibitem{taylor2015causal}
Brian Taylor, Vasiliy Karasev, and Stefano Soatto.
\newblock Causal video object segmentation from persistence of occlusions.
\newblock In {\em CVPR}, 2015.

\bibitem{tokmakov2017learning}
Pavel Tokmakov, Karteek Alahari, and Cordelia Schmid.
\newblock Learning motion patterns in videos.
\newblock In {\em CVPR}, 2017.

\bibitem{tokmakov2018learning}
Pavel Tokmakov, Cordelia Schmid, and Karteek Alahari.
\newblock Learning to segment moving objects.
\newblock {\em IJCV}, Sep 2018.

\bibitem{DetectronPytorch}
Roy Tseng.
\newblock Detectron.pytorch.
\newblock \url{https://github.com/roytseng-tw/Detectron.pytorch}.

\bibitem{wang2015saliency}
Wenguan Wang, Jianbing Shen, and Fatih Porikli.
\newblock Saliency-aware geodesic video object segmentation.
\newblock In {\em CVPR}, 2015.

\bibitem{wang2019learning}
Wenguan Wang, Hongmei Song, Shuyang Zhao, Jianbing Shen, Sanyuan Zhao,
  Steven~CH Hoi, and Haibin Ling.
\newblock Learning unsupervised video object segmentation through visual
  attention.
\newblock In {\em CVPR}, 2019.

\bibitem{wertheimer1923untersuchungen}
Max Wertheimer.
\newblock Untersuchungen zur lehre von der gestalt. ii.
\newblock {\em Psychologische forschung}, 4(1):301--350, 1923.

\bibitem{xie2018object}
Christopher Xie, Yu Xiang, Dieter Fox, and Zaid Harchaoui.
\newblock Object discovery in videos as foreground motion clustering.
\newblock 2019.

\bibitem{xu2018youtube}
Ning Xu, Linjie Yang, Yuchen Fan, Dingcheng Yue, Yuchen Liang, Jianchao Yang,
  and Thomas Huang.
\newblock {YouTube-VOS}: A large-scale video object segmentation benchmark.
\newblock In {\em ECCV}, 2018.

\bibitem{yang2015self}
Yanchao Yang, Ganesh Sundaramoorthi, and Stefano Soatto.
\newblock Self-occlusions and disocclusions in causal video object
  segmentation.
\newblock In {\em ICCV}, 2015.

\end{thebibliography}
}

\clearpage

\ificcvfinal{
    \iftoggle{includeappendix}{\section{Appendix}
We include an additional comparison to prior methods on DAVIS'16 motion
segmentation in \Cref{sec:davis16-moving}, to a prior method on FBMS in
\Cref{sec:fbms_moving}, and further implementation details in
\Cref{sec:implementation}.

\subsection{DAVIS'16 Motion Segmentation}
\label{sec:davis16-moving}
We compare our method with motion foreground-background segmentation methods on DAVIS 2016. We convert our instance segmentation output into binary motion masks by marking as foreground any pixel belonging to a predicted instance with a score $>0.7$, and report results in \Cref{tab:davis16-final}. Although this is not our target task, our method compares favorably to the state of the art, and even modestly improves the F-measure for boundary accuracy ($\mathcal{F}$) by 1\%. On the $\mathcal{J}$ metric it is only outperformed by a concurrent work by 0.7\%.

\subsection{FBMS Moving Only}\label{sec:fbms_moving}
A recent line of work~\cite{bideau2016detailed,bideau2018best} proposed
evaluating a subtask of spatiotemporal grouping. Whereas standard spatiotemporal
grouping requires segmenting and tracking all instances that move at any point
in the video,~\cite{bideau2018best} focuses on segmenting and tracking instances
\textit{only} in frames where they move.

In order to evaluate this subtask,~\cite{bideau2018best} uses an alternative
labeling for FBMS introduced in~\cite{bideau2016detailed}, and
supplements the official FBMS measure with a $\Delta$ Obj metric, which
indicates the average absolute difference between the number of predicted
objects and groundtruth objects in each sequence. Intuitively, this
penalizes false-positive detection of static objects or background; we refer
the reader to~\cite{bideau2018best} for further information. As with the
official FBMS metric, the precision, recall and F-measure do not penalize static
detections.

As described in Sec. 3.4 of our main submission, our final approach
uses the appearance stream to track objects even after they stop
moving. For a fair comparison, we disable this component, applying tracking
directly to the output of our two-stream model that detects only moving objects
in each frame.

\Cref{tab:their-measure-their-labels} shows that our method significantly
reduces the number of false positive segmentations compared to both~\cite{xie2018object,bideau2018best} as evidenced by the
improvement in $\Delta$ Obj of nearly 65\% (from 4 to 1.4), while performing competitively with~\cite{xie2018object} on F-measure.

\begin{table}[t]
\centering
\resizebox{\linewidth}{!}{%
\begin{tabular}{@{}llr@{\hspace{0.9em}}r@{\hspace{0.9em}}r@{\hspace{0.9em}}r@{\hspace{0.9em}}r}
\toprule
\multicolumn{2}{c}{Measure}
& LSMO~\cite{tokmakov2018learning} & MotAdapt \cite{siam2018video} & AGS \cite{wang2019learning} & Ours \\
\midrule
\multirow{3}{*}{$\mathcal{J}$}
& Mean   &  78.2 & 77.2 & \textbf{79.7} & 79.0 \\
& Recall &  89.1 & 87.8 & 91.1 & \textbf{92.7}  \\
& Decay  &   4.1 & 5.0 &  \textbf{1.9} & 3.7 \\
\addlinespace{}
\multirow{3}{*}{$\mathcal{F}$}
& Mean   &  75.9 & 77.4 & 77.4 & \textbf{78.4} \\
& Recall &  84.7 & 84.4 & 85.8 & \textbf{86.7}  \\
& Decay  &   3.5 &  3.3 & \textbf{1.6} & 5.4 \\
\addlinespace{}
$\mathcal{T}$
& Mean   & \textbf{21.2} & 27.9 & 26.7 & 25.2 \\
\bottomrule
\end{tabular}
}
\caption{DAVIS '16 results on the validation set using the intersection over
union ($\mathcal{J}$), F-measure ($\mathcal{F}$), and temporal stability
($\mathcal{T}$) metrics.}
\label{tab:davis16-final}
\end{table}

\begin{table}
\footnotesize
    \centering
    \begin{tabular}{lccccc}
      \toprule
        & P           & R           & F           & $\Delta$ Obj\\\midrule
    \cite{xie2018object} & 75.9 & 66.6 & \textbf{67.3} & 4.9 \\
    \cite{bideau2018best}
        & 74.2        & 63.1  & 65.0        & 4 \\
    Ours-J
        & 77.1  & 62.6        & 66.3  & {\bf 1.4} \\\bottomrule
    \end{tabular}
    \caption{FBMS TestSet results with alternative evaluation
    from~\cite{bideau2016detailed,bideau2018best}: precision (P), recall (R),
    F-measure (F), and difference in number of predicted and groundtruth objects
    ($\Delta$ Obj, lower is better).}
    \label{tab:their-measure-their-labels}
\end{table}

\subsection{Failure cases}
\label{sec:failure}
We also present illustrative failure cases of our method in
\Cref{fig:qualitative-mistakes}. The most common mistakes our method makes are
tracking failures and the misclassification of static objects as moving. In the
top row, the furthest horse (colored purple in the first frame) is completely
occluded by the man in the middle frame, leading to an identity swap in the last
frame, indicated by the color swap from purple to yellow. In the bottom row, the
car further to the right is parked and not moving but is classified as moving by
our approach. Our hypothesis, based on viewing similar failure cases, is that
this is due to the proximity of the static car to the moving van, which our
learning-based approach may have learnt to use as a cue for identifying moving
objects.

\begin{figure}
    \centering
    \includegraphics[width=\linewidth]{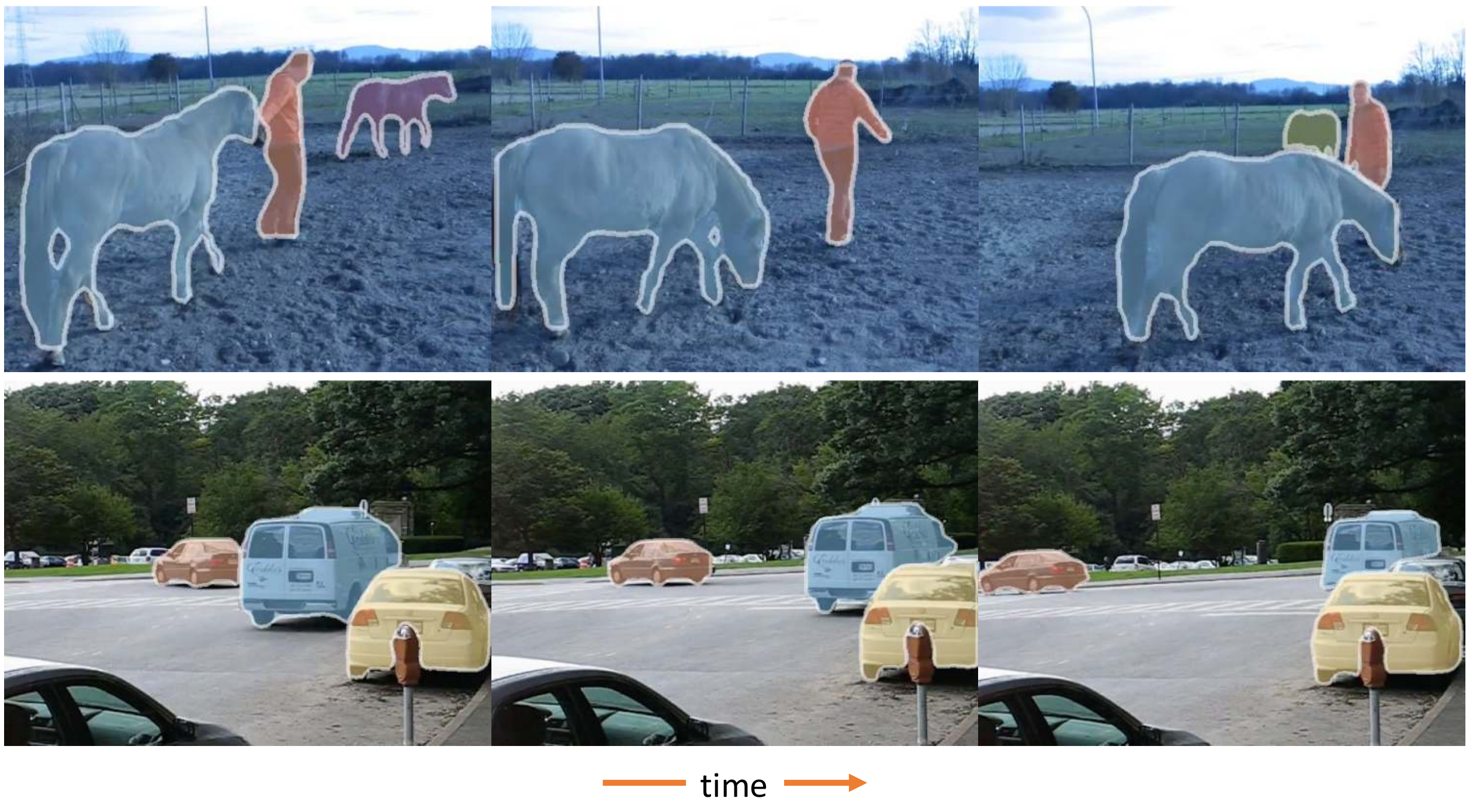}
    \caption{Illustrative failure cases of our method. The most common failures are due to tracking or mis-classification of static objects near the camera as moving objects. Top: Our overlap-based tracker fails due to complete occlusions, such as the occlusion of the horse by the person in the middle frame. Bottom: Static objects near the camera are occasionally mistakenly detected as moving objects, such as the white car on the right.}
\label{fig:qualitative-mistakes}
\end{figure}

\subsection{Implementation details}\label{sec:implementation}

\smallsec{Miscellaneous} We encode optical flow following~\cite{tokmakov2018learning}: We use a 3-channel image for ease of use with image-based CNNs, where the first channel encodes the angle and the second channel encodes the magnitude at each pixel, and the last channel is empty. For extracting flow, we use the version of FlowNet2 trained on synthetic data and fine-tuned on real data. For all visualizations throughout this paper, we use a confidence threshold of $0.7$.

\smallsec{Training Regime} We train the appearance stream on COCO~\cite{lin2014microsoft} to convergence (90,000 iterations). We train our motion stream on FlyingThings3D using Groundtruth flow as input, and then with FlowNet2 flow~\cite{ilg2017flownet} for 10,000 iterations each. We train the joint model on YTVOS and then on DAVIS for 5,000 iterations each.}{}
}

\end{document}